\documentclass[]{style/style}

\usepackage{microtype}
\usepackage{graphicx}
\usepackage{subcaption}
\usepackage{csquotes}
\usepackage{afterpage}
\usepackage{xcolor}

\usepackage{amsmath}
\usepackage{amssymb}
\usepackage{mathtools}
\usepackage{amsthm}
\usepackage{xspace}
\usepackage{colortbl}
\usepackage{multirow}
\usepackage{multicol}
\usepackage{enumitem}
\usepackage{wrapfig}
\usepackage{nicefrac}
\usepackage[font=small,labelfont=bf,justification=justified,singlelinecheck=false]{caption}

\usepackage{booktabs}
\usepackage{multirow}
\usepackage{graphicx}
\usepackage{xcolor}
\usepackage{geometry}
\usepackage{array}
\usepackage{tabularx} 
\newcolumntype{L}[1]{>{\raggedright\arraybackslash}p{#1}}
\newcolumntype{Y}{>{\raggedright\arraybackslash}X}

\usepackage{amsmath}  
\newcommand{\model}{Wan-Streamer\xspace}

\title{Wan-Streamer v0.1: End-to-end Real-time Interactive Foundation Models}

\author[]{Wan Team, Alibaba Group}

\affiliation[]{\small See \hyperref[app:contributions-and-acknowledgements]{Contributions and Acknowledgements} for the full author list.}

\abstract{
We present \textbf{Wan-Streamer}, a native-streaming, end-to-end interactive foundation model designed from the ground up for real-time, low-latency, full-duplex audio-visual interaction. Wan-Streamer seamlessly models language, audio, and video as both input and output within \textbf{a single Transformer}, where the sequence is represented as interleaved visual, audio, and text input tokens together with visual, audio, and text output tokens, coordinated by block-causal attention for incremental streaming. Unlike cascaded interactive systems that rely on separate VAD, ASR, language, TTS, audio-driven animation, or video-generation modules, Wan-Streamer does not rely on external language, speech, avatar, or video-generation modules: perception, reasoning, generation, response timing, turn management, and cross-modal synchronization are learned jointly within one unified model, reducing pipeline latency and error accumulation. To support natural audio-visual responsiveness, we redesign the entire stack around streamability, including causal encoders, causal decoders, block-causal attention, and low-latency multimodal token scheduling, enabling streaming units as short as 160 ms at 25 fps. Wan-Streamer achieves approximately \textbf{200 ms} model-side response latency and approximately 550 ms total interaction latency when combined with 350 ms bidirectional network latency, supporting sub-second duplex audio-visual communication. These results position Wan-Streamer as a unified, end-to-end, multimodal interactive foundation model for low-latency streaming interaction.
}

\metadata[Website]{\url{https://wan-streamer.com/}}

\begin{document}

\maketitle

\section{Introduction}
\label{section:intro}

Human interaction with the physical world is fundamentally streaming and full-duplex. People do not first finish perceiving, then reason in isolation, and only afterwards produce a response. Instead, they continuously watch, listen, speak, gesture, react, pause, and interrupt, with perception and expression overlapping at audio-visual timescales. Building artificial systems with the same interaction pattern is becoming increasingly important for embodied assistants, real-time digital humans, live broadcasting, interactive entertainment, and world models that can be explored or controlled online~\cite{lingbotva2026,matrixgame2026,ding2025kling-avatar}. These applications require more than a model that can understand an image, generate a clip, or answer a text prompt. They require a real-time interactive foundation model: a model that continuously consumes audio-visual observations, maintains a persistent world and dialogue state, decides when and how to respond, and expresses that response through synchronized language, speech, and video with very low latency.

Recent progress has advanced several pieces of this goal. Multimodal language models can reason over visual and acoustic inputs~\cite{bai2025qwen2.5vl,team2025kimi-vl,hong2025glm4.5v}, while large-scale video generation models can synthesize increasingly realistic motion and appearance from text, audio, camera trajectories, or actions~\cite{wan2025wan,seedance2026seedance,yang2025cogvideox,ding2025kling-avatar,bai2025recammaster,matrixgame2026}. Causal and streaming generation methods further make incremental synthesis feasible by replacing offline bidirectional generation with cached context and rolling prediction~\citep{huang2025selfforcing,chen2024diffusionforcing,talkingmachines2025}. However, these advances are still usually assembled as asymmetric or cascaded systems. Some systems perceive audio and video but respond only in text or speech; others generate audio-visual behavior but rely on external language, ASR, TTS, animation, or rendering modules. Even when the interface appears multimodal, text is often used as a hidden intermediate representation between separately trained components~\cite{mavid2025,mio2025}. Such pipelines introduce waiting time at module boundaries, accumulate recognition and synchronization errors, and make response timing, turn management, identity preservation, and long-horizon consistency difficult to learn as part of one behavior.

The core difficulty is that real-time audio-visual interaction is not simply the union of multimodal understanding and multimodal generation. It is intrinsically full-duplex rather than a simple alternation between input and output: when the user is speaking, the agent should still produce visible listening behavior, and when the agent is responding, it should still perceive the user's audio-visual feedback for interruption and adaptation. Different modalities have different token rates, representations, objectives, and latency constraints, yet they must be causally aligned within a single ongoing process. Incoming speech and video should immediately affect outgoing speech and motion; generated audio and visual states should be coupled before decoding, not repaired afterwards; and every emitted unit should become part of the full interaction history so that the model can preserve identity, scene state, speaking rhythm, and user intent over long sessions. These requirements make streamability a modeling constraint rather than a serving optimization. A system designed around offline encoders, bidirectional video decoders, round-based dialogue, or post-hoc audio-visual synchronization cannot recover truly low-latency full-duplex behavior by engineering alone.

\begin{figure}[t]
    \centering
    \includegraphics[width=\linewidth]{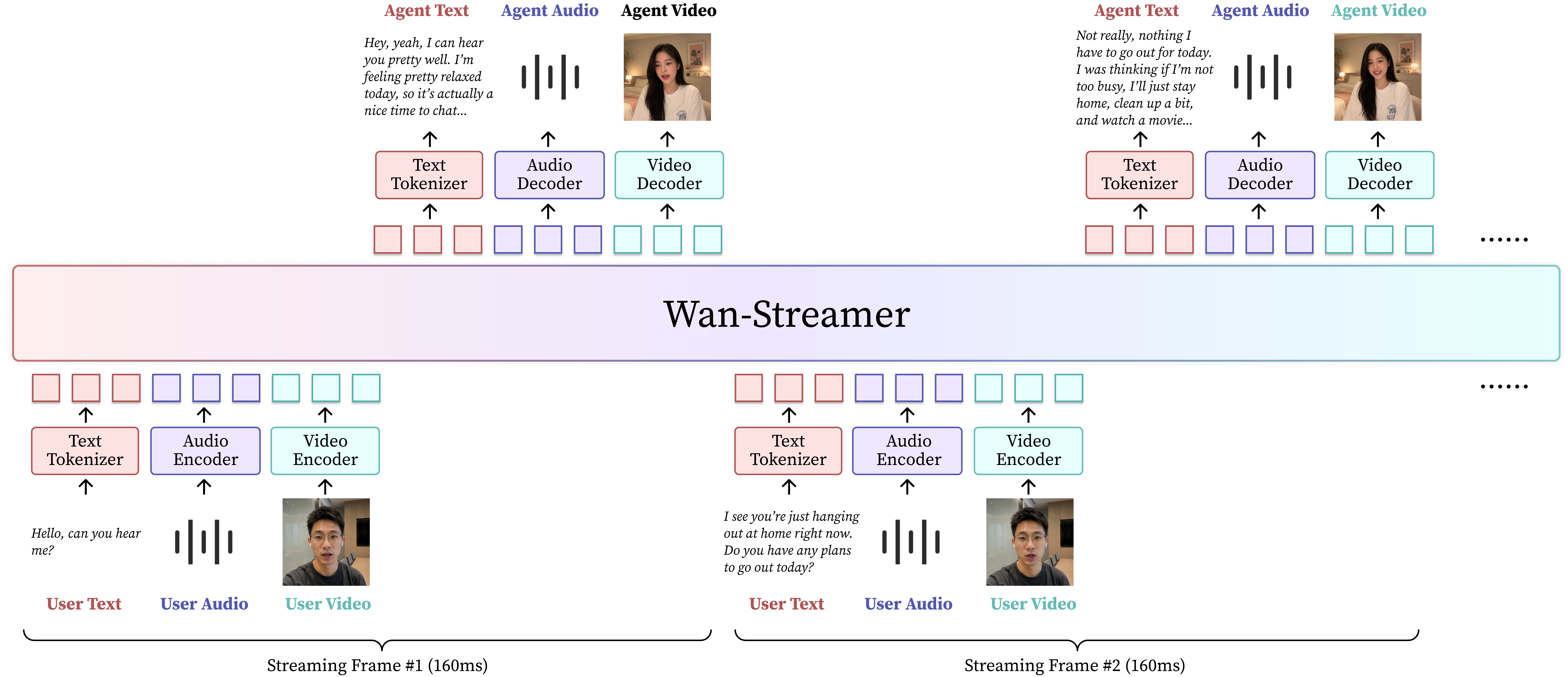}
    \caption{Overview of Wan-Streamer. It models language, audio, and video as both input and output within \textbf{a single Transformer}, using block-causal attention for incremental streaming generation.}
    \label{fig:framework}
\end{figure}

To address this challenge, we design \textbf{Wan-Streamer} from the ground up as a native-streaming, end-to-end model for real-time full-duplex audio-visual interaction, as illustrated in Fig.~\ref{fig:framework}. Wan-Streamer is built around one streaming contract: every component must operate causally, every newly observed unit must be usable immediately, and every generated unit must be emitted and committed back into the interaction history. Wan-Streamer represents language, audio, and video, on both the input and output sides, as an interleaved causal sequence processed by \textbf{a single Transformer}. The model does not rely on external VAD, ASR, language, TTS, audio-driven animation, or video-generation modules. Instead, audio-visual perception, semantic reasoning, response planning, speech generation, visual generation, response timing, and turn-taking behavior are optimized jointly within one persistent interaction state. To make this possible, the entire stack is designed for causality from the beginning: strictly causal audio and video variational autoencoders (VAEs) for streaming latent coding, causal audio-visual encoders, causal audio and video decoders, and a temporally causal Transformer coordinated by block-causal attention.

During inference, Wan-Streamer remains a single end-to-end model, but we deploy it as a thinker-performer streaming pipeline to maximize overlap and hardware utilization. At streaming step $k$, the thinker first consumes the current user audio-visual observations, applies the causal audio-visual encoders, and runs the short token-causal Transformer pass for language prediction and state update. This produces the new KV-cache slice for the current interaction state. At the communication boundary, the thinker receives the audio and video latents generated by the performer for the previous response unit, and sends the current KV-cache slice to the performer. The thinker then decodes those previous latents into output audio and video for immediate emission, while the performer uses the newly received full-history KV context to run only the flow-matching solver for the next audio-visual latent unit. The resulting latents are kept on the performer and returned to the thinker at the next streaming step. This schedule preserves the unified causal state through KV exchange, but places expensive latent generation on the performer and allows current-frame perception/state update, previous-frame output decoding, next-frame latent denoising, and KV/latent communication to overlap across adjacent streaming units. Together with CUDA graph capture, compilation, and optimized kernels, Wan-Streamer reaches approximately \textbf{200 ms} model-side response latency. With 350 ms bidirectional network latency, the total interaction latency is approximately 550 ms, supporting sub-second audio-visual communication without the module-boundary waiting time of typical cascaded real-time dialogue systems.

Our contributions are summarized as follows:
\begin{itemize}
    \item We introduce Wan-Streamer, a native-streaming, end-to-end interactive foundation model that supports language, audio, and video as both inputs and outputs within a single Transformer, without relying on external language, speech, animation, or video-generation modules.
    \item We develop a fully causal multimodal architecture for real-time interaction, including strictly causal audio and video VAEs, causal audio-visual encoders and decoders, block-causal multimodal attention, and full-history autoregressive streaming.
    \item We present a low-latency thinker-performer inference system that preserves the unified model state through KV-cache exchange while overlapping understanding and generation, achieving approximately 200 ms model-side response latency and approximately 550 ms total interaction latency.
\end{itemize}

\FloatBarrier

\section{Method}
\label{sec:method}

\subsection{Overview}
Wan-Streamer models interaction as a continuous causal stream in which user observations and agent responses jointly update the ongoing context. At the $k$-th streaming unit, let
$u_k=(u_k^{\mathrm{t}},u_k^{\mathrm{a}},u_k^{\mathrm{v}})$ denote the user's language, audio, and video observations, and let
$y_k=(y_k^{\mathrm{t}},y_k^{\mathrm{a}},y_k^{\mathrm{v}})$ denote the agent response. The model encodes the currently available user observations and predicts the next response from the complete causal history across both sides of the interaction:
\begin{equation}
\label{eq:ar-factorization}
    p_\theta(y_{1:K}\mid u_{1:K})=\prod_{k=1}^{K} p_\theta\!\left( y_k^{\mathrm{t}},y_k^{\mathrm{a}},y_k^{\mathrm{v}} \mid u_{\le k}^{\mathrm{t}},u_{\le k}^{\mathrm{a}},u_{\le k}^{\mathrm{v}}, y_{<k}^{\mathrm{t}},y_{<k}^{\mathrm{a}},y_{<k}^{\mathrm{v}} \right),
\end{equation}
where $K$ denotes the number of streaming units. Once generated, the response unit is appended together with the corresponding user observations to the history state and becomes context for the next unit. The language response is represented as a sequence of discrete tokens and optimized with next-token prediction using cross-entropy loss.
The audio and video responses are represented in continuous latent spaces and generated jointly with conditional flow matching. For the current response unit, clean states and noisy states play different roles. For modality $m\in\{\mathrm{a},\mathrm{v}\}$, let $z_0^m$ be the clean target latent of the response, and let $\epsilon^m\sim\mathcal N(0,I)$ be Gaussian noise. At flow time $\tau$, we construct a noisy latent $z_\tau^m$ using
\begin{equation}
\label{eq:flow-path}
    z_\tau^m=(1-\tau)z_0^m+\tau\epsilon^m, \qquad 
    \frac{\partial z_\tau^m}{\partial\tau} = \epsilon^m-z_0^m.
\end{equation}
Let $c_k=\{u_{\le k}^{\mathrm{t}},u_{\le k}^{\mathrm{a}},u_{\le k}^{\mathrm{v}},y_{<k}^{\mathrm{t}},y_{<k}^{\mathrm{a}},y_{<k}^{\mathrm{v}}\}$ denote the clean streaming context, consisting of user observations that have arrived and agent responses that have already been committed to history. The current audio and video responses are the noisy variables being denoised, while the context remains available as causal history. We train the model to estimate the velocity fields for both noisy audio and video latents conditioned on $c_k$ and noise level $\tau$:
\begin{equation}
\label{eq:joint-flow-matching}
    \mathcal L_{\mathrm{FM}}^m =  \mathbb{E}_{\epsilon^m} \left\| f_\theta \!\left(z_\tau^{\mathrm{a}},z_\tau^{\mathrm{v}},c_k,\tau\right) - \frac{\partial z_\tau^m}{\partial\tau} \right\|_2^2 ,
\end{equation}
where $f_\theta$ denotes the unified diffusion transformer. The same clean context conditions both velocity predictions, allowing speech, motion, appearance, and scene evolution to be optimized as a coupled response. After denoising, the estimated clean latents are appended directly to the history as clean context for subsequent streaming units, while the causal decoders render them into external audio and video outputs.


\subsection{Data}
Wan-Streamer is trained on a broad mixture of understanding, generation, and end-to-end interaction data. For understanding-oriented learning, we include image, audio, and video understanding data, text dialogue, ASR, TTS, audio dialogue, and related language-audio-visual supervision. These data teach the model to convert streaming multimodal observations into a shared causal context and to preserve dialogue competence under multimodal inputs. For generation-oriented learning, we include image generation, audio generation, video generation, and joint audio-visual generation tasks, covering both single-modality and cross-modality conditions. Finally, we use end-to-end duplex interaction data in which text, audio, and video can appear on both the input side and the output side. This end-to-end data component exposes the model to the target setting of simultaneous multimodal perception and expression.

\subsection{Training}
Training proceeds in three stages. The first stage is independent-task pretraining. We initialize the unified Transformer from a language model~\citep{yang2025qwen25,yang2025qwen3} and train the multimodal interface around it with the mixture described above. On the understanding side, the causal audio and video encoders are trained together with the Transformer so that image, audio, video, and dialogue understanding metrics approach those of dedicated multimodal understanding models, while conversational ability remains comparable to turn-based dialogue models of similar scale. On the generation side, the same Transformer is trained with the causal audio and video latent spaces on image, audio, video, and joint audio-visual generation tasks. These understanding and generation tasks are mixed during pretraining so that perception, language reasoning, and latent generation are aligned in one sequence model rather than optimized as isolated modules.

The second stage is end-to-end interaction training. We train on duplex interaction data where user text/audio/video inputs and agent text/audio/video outputs are interleaved in the same causal stream. This stage adapts the pretrained model from independent tasks to the target real-time setting: the model must update its state from current user observations, generate synchronized language, audio, and video responses, and commit the generated clean latents back into history for subsequent streaming units. As a result, response timing, active listening behavior, interruption handling, and long-context consistency are learned under the same causal format used at inference time.

The third stage is distillation for low-latency streaming. A stronger teacher with classifier-free guidance (CFG) and more flow-matching solver steps is distilled into an efficient student used at deployment. This distillation absorbs the effect of CFG into the student and reduces the number of solver steps while preserving audio-visual quality. We also use rolling distillation to mitigate long-horizon degradation: the student is rolled out over consecutive streaming units and trained on its own generated history, using a self-forcing strategy~\citep{huang2025selfforcing} with distribution matching~\citep{yin2024onestepdiffusiondistributionmatching,yin2024improveddistributionmatchingdistillation} to align the student trajectory with the teacher under realistic rollout conditions. This substantially reduces train-test mismatch and improves long-form generation quality.

\subsection{Inference}
\label{sec:inference}
Although Wan-Streamer is trained as a single end-to-end model, we deploy it as a separated thinker-performer pipeline to maximize overlap and hardware utilization. The \textit{thinker} hosts the causal audio and video encoders, the short token-causal Transformer path for language prediction and state update, KV-cache construction, and the causal audio and video decoders. The \textit{performer} hosts only the latent generation path. After system prefill, the thinker broadcasts the initial KV cache to the performer so that both sides share the same full-history state.

At streaming step $k$, the thinker consumes the current user audio-visual observations, applies the causal encoders, and runs token-causal decoding over the language and state slots to produce the current KV-cache slice. Around the same communication boundary, the thinker receives from the performer the clean audio and video latents produced in the preceding step, sends the newly produced KV slice to the performer, and decodes the returned latents into the audio-visual output that is emitted immediately. The performer appends the received KV slice into its own full-history cache and runs the flow-matching solver only for the next audio-visual latent unit. The resulting clean latents remain on the performer and are sent back to the thinker at the next streaming step for decoding and emission.

\begin{figure}[t]
    \centering
    \includegraphics[width=\linewidth]{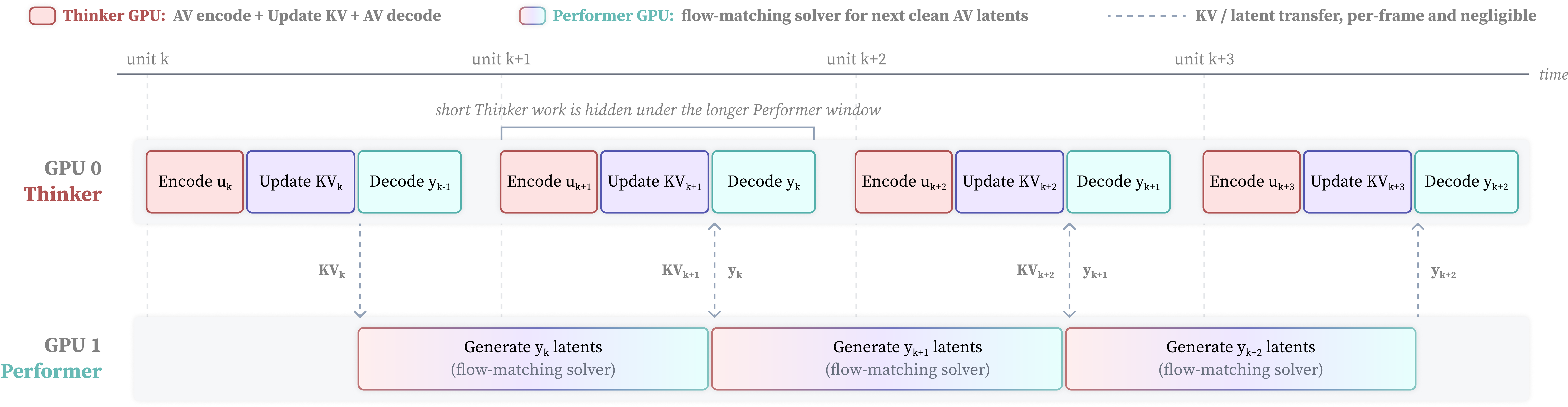}
    \caption{Thinker-performer overlap during streaming inference. At unit $k$, the thinker encodes the current user observations $u_k$, updates the KV cache, and decodes the previous response latents $y_{k-1}$ for immediate emission. The performer receives the current KV slice and runs only the flow-matching solver to generate the next clean audio-visual latents $y_k$, which are returned to the thinker at the following unit. For clarity, performer windows are drawn as one full 160 ms streaming unit; in practice, real-time throughput only requires the performer time plus KV/latent communication to be below this unit duration.}
    \label{fig:thinker-performer-overlap}
\end{figure}

As shown in Fig.~\ref{fig:thinker-performer-overlap}, this schedule pipelines current-frame perception and state update, previous-frame audio-visual decoding, KV/latent communication, and next-frame latent denoising across adjacent streaming units. After overlap, per-frame throughput is determined mainly by the performer wall time: the system can run in real time as long as the performer time, plus the small KV-cache and latent communication overhead, fits within one 160 ms streaming unit. This throughput condition is distinct from model-side response latency, which measures the full signal-to-signal path from receiving a user unit to emitting the corresponding response unit. That latency is the sum of encoding, thinker state update, performer latent generation, and decoding, and is currently approximately \textbf{200 ms}. Since the performer does not run decoders and the thinker does not run the expensive flow-matching solver, the deployment preserves the semantics of one unified model while overlapping most latency-critical work. In practice, we further use CUDA graph capture, compilation, optimized kernels, and KV-cache exchange to improve throughput.

\FloatBarrier

\section{Experiments}

\textbf{Latency and runtime comparison.}
For real-time interaction, the most relevant latency is the delay from the user's latest signal to the first perceptible assistant response. For \model, we measure the two-GPU thinker-performer serving path described in Sec.~\ref{sec:inference}. Model-side signal-to-signal latency starts when a 160 ms user streaming unit is available to the thinker and ends when the corresponding audio-video response unit has been decoded for emission at 25 FPS. Under this protocol, \model reaches about 200 ms model-side latency; adding a 350 ms bidirectional network budget gives about 550 ms total interaction latency for a remote user.

Public systems do not always expose the same endpoint. Tab.~\ref{tab:response-latency-comparison} compares speech and omni-modal dialogue systems by the closest available user-visible response latency, while keeping model-only, first-packet, and API/product measurements separate when they are not directly comparable.

\begin{table}[!t]
    \caption{Response-latency comparison for real-time speech and omni-modal interaction systems. The shaded row highlights \model; ``N/R'' means no aligned absolute response latency is publicly reported.}
    \label{tab:response-latency-comparison}
    \centering
    \scriptsize
    \setlength{\tabcolsep}{3pt}
    \renewcommand{\arraystretch}{1.04}
    \begin{tabularx}{\textwidth}{@{}L{.16\textwidth}L{.15\textwidth}L{.20\textwidth}L{.22\textwidth}Y@{}}
        \toprule
        System & Interaction & User-visible response & Other reported metric & Comparison boundary \\
        \midrule
        Doubao Realtime Voice~\citep{doubao_realtime_voice2025,volcengine_realtime_voice2025} & speech-to-speech & $\sim$1 s overall & $\sim$700 ms bare-model latency & Official speech-only product numbers; no visual agent output. \\
        Seeduplex~\citep{seeduplex2026} & speech-to-speech & N/R absolute & $-$250 ms endpoint, $-$300 ms interruption latency vs. previous Doubao & Relative production improvement; speech-only. \\
        GPT-4o / Realtime API~\citep{openai2024gpt4o,latent2024realtimeapi,hume2025evi3} & speech-to-speech, audio/vision input & protocol-dependent & 232/320 ms official audio response; $\sim$500 ms API TTFB; $\sim$800 ms target voice-to-voice & Reported numbers mix model response, API TTFB, endpointing, and network. \\
        Hume EVI 3~\citep{hume2025evi3} & speech-to-speech & 0.9--1.4 s web-app benchmark & under 300 ms model response & Vendor benchmark; no visual output stream. \\
        Gemini Live API~\citep{hume2025evi3} & speech-to-speech & 1.2--3.6 s API benchmark & N/R model-side & Vendor benchmark; not an official model breakdown. \\
        Sesame web app~\citep{hume2025evi3} & speech-to-speech & 0.8--1.2 s web-app benchmark & N/R model-side & Vendor benchmark; speech-only. \\
        Moshi~\citep{defossez2024moshi} & speech-to-speech & N/R product path & 160 ms theoretical; 200 ms practical model latency & Native full-duplex speech model; no visual agent. \\
        Qwen3/3.5-Omni~\citep{qwen3omni2025,qwen35omni2026} & audio-video-text in, speech/text out & N/R interaction loop & first-packet: 234/547 ms; Qwen3.5 Flash 235/426 ms, Plus 435/651 ms & First-packet metric; no synchronized visual avatar generation. \\
        MiniCPM-o 4.5~\citep{minicpmo45} & audio-video in, speech/text out & N/R interaction loop & 0.58 s first-token; RTF 0.20--0.27 & First-token/RTF metric; no visual avatar generation. \\
        \specialrule{0.35pt}{.35em}{.15em}
        \rowcolor{stylebg}
        \textbf{\model (ours)} & text/audio/video in/out & \textbf{$\sim$550 ms total} including 350 ms network & \textbf{$\sim$200 ms model-side}; 25 FPS video output & One end-to-end model; text I/O, speech, and synchronized visual response share one causal stream. \\
        \bottomrule
    \end{tabularx}
\end{table}
\FloatBarrier

Tab.~\ref{tab:response-latency-comparison} should be read by measurement boundary rather than by the smallest raw number. Some public systems report model-internal, first-packet, first-token, endpointing, or API time-to-first-byte latency, which is useful for engineering but does not always correspond to the delay perceived by a remote user. Several omni-modal systems further accept audio or video input but do not close the loop with synchronized visual agent output. We therefore report both the $\sim$200 ms model-side streaming latency and the $\sim$550 ms total interaction latency for \model, and keep these numbers separate from partial or speech-only measurements.

For visually embodied systems, the available measurements are less uniform. Some systems are full interaction loops, while others are renderers or audio-visual generators driven by external dialogue and speech modules. Tab.~\ref{tab:av-runtime-comparison} therefore records the runtime metric reported by each source, such as FPS, first-frame delay, chunk latency, or audio-to-visual delay, together with the part of the interaction stack that the system covers.

To make this comparison readable, we separate full-loop or interactive digital-human systems from avatar rendering and joint audio-visual generation components. The first group is closer in interaction scope, but often reports only real-time operation, FPS, or qualitative bounded-latency claims instead of absolute signal-to-signal response latency. The second group usually provides clearer rendering-side metrics, but assumes external dialogue, speech, or perception modules. This is why Tab.~\ref{tab:av-runtime-comparison} reports both runtime and covered scope, rather than treating every FPS, first-frame, or audio-to-visual number as an end-to-end interaction latency.

Under this convention, the key comparison is whether the reported path includes user perception, response timing, speech generation, and synchronized visual output. A renderer can be fast once clean audio or text is available, but its perceived interaction latency still depends on the upstream dialogue and speech stack. Conversely, an interactive system can claim real-time operation while leaving the latency boundary unspecified. \model is evaluated against both axes: it reports the full remote audio-visual response path and also specifies the model-side streaming runtime.

Taken together, these comparisons emphasize why raw speed alone is insufficient. Speech-only systems can report very low model or first-packet latency, but they do not generate a synchronized visual response. Avatar and audio-visual generation systems can run at 20--40 FPS, but many rely on external dialogue, speech, or perception modules, and their published runtime does not include the whole conversational path. \model combines these parts in one causal stream: text I/O, user audio-video perception, response timing, speech generation, and 25 FPS visual expression are produced by the same end-to-end model. Thus, its 550 ms total latency covers the full audio-visual response path.

\begin{table}[!t]
    \caption{Runtime comparison with visual agents, streaming avatars, and audio-visual generators. Most numbers here are component-level runtime metrics rather than the aligned response latency in Tab.~\ref{tab:response-latency-comparison}; the shaded row highlights \model.}
    \label{tab:av-runtime-comparison}
    \centering
    \scriptsize
    \setlength{\tabcolsep}{3pt}
    \renewcommand{\arraystretch}{1.04}
    \begin{tabularx}{\textwidth}{@{}L{.18\textwidth}L{.25\textwidth}L{.23\textwidth}Y@{}}
        \toprule
        System & Visual interaction scope & Reported runtime & Main difference from \model \\
        \midrule
        \multicolumn{4}{@{}l}{\emph{Full-loop or interactive digital-human systems}} \\
        Body of Her~\citep{ao2024bodyofher} & end-to-end humanoid agent & next frame within 42 ms at 24 FPS & Preliminary unified agent; no deployed signal-to-signal latency. \\
        MIDAS~\citep{chen2025midas} & multimodal digital-human video synthesis & real-time frame-by-frame generation & Does not disclose absolute response latency. \\
        U-Mind~\citep{deng2026umind} & text, speech, motion, and video interaction loop & real-time video rendering claimed & Text-first pipeline; latency breakdown not public. \\
        X-Streamer~\citep{xstreamer2025} & open-ended video chat from a portrait & 25 FPS multimodal streaming on two A100 GPUs & Absolute response latency is not disclosed. \\
        LPM 1.0~\citep{lpm2026} & online character performance engine & low-latency real-time causal streaming & Visual engine coupled to external A2A systems; latency is not intrinsic to LPM alone. \\
        MAViD~\citep{mavid2025} & audio-visual dialogue framework & no absolute latency reported & Modular framework; useful for capability comparison, not latency comparison. \\
        M.I.O~\citep{mio2025} & interactive omni-avatar system & bounded-latency design discussed & Multi-module embodied system; no public signal-to-signal number. \\
        \addlinespace[.25em]
        \multicolumn{4}{@{}l}{\emph{Avatar rendering or joint audio-visual generation components}} \\
        VASA-1~\citep{xu2024vasa} & audio-driven talking face & 40 FPS with 170 ms preceding latency & Renderer only; no dialogue reasoning or user visual perception. \\
        TalkingMachines~\citep{talkingmachines2025} & FaceTime-style audio-driven video & real-time chunk generation by TTBC & Relies on an external audio LLM for dialogue and speech. \\
        StreamAvatar~\citep{streamavatar2025} & streaming talking/listening avatar & FFD 0.33--0.39 s; video latency $\sim$1.20 s & Avatar renderer driven by speech/audio; no unified dialogue model. \\
        Avatar Forcing (Ki et al.)~\citep{ki2026avatarforcing} & interactive head-avatar reactions & $\sim$500 ms reaction latency; 6.8$\times$ speedup & Reacts to user audio/motion, but does not generate dialogue speech. \\
        AvatarForcing (Cui et al.)~\citep{cui2026avatarforcing} & one-step streaming talking avatar & 34 ms/frame; 0.51 s audio-to-visual delay & Strong visual streaming metric, not perceptual dialogue. \\
        LiveTalk~\citep{livetalk2025} & multimodal interactive avatar video & 24.82 FPS; 0.33 s first-frame latency & Uses Qwen3-Omni for speech reasoning; video latency is separate. \\
        Hallo-Live~\citep{hallolive2026} & text-driven joint audio-video avatar & 20.38 FPS with 0.94 s latency & Text-driven; does not continuously perceive user audio-video. \\
        OmniForcing~\citep{omniforcing2026} & text-to-audio-video streaming generation & TTFC $\sim$0.7 s; $\sim$25 FPS & First-chunk generation latency, not user response latency. \\
        \specialrule{0.35pt}{.35em}{.15em}
        \rowcolor{stylebg}
        \textbf{\model (ours)} & text/audio/video perceptual dialogue with synchronized speech and video output & \textbf{25 FPS}; \textbf{$\sim$550 ms total}; \textbf{$\sim$200 ms model-side} & Single causal Transformer learns text I/O, perception, speaking, listening behavior, interruption, and visual response together. \\
        \bottomrule
    \end{tabularx}
\end{table}
\FloatBarrier

\textbf{Naturalness.}
Beyond response speed, \model improves interaction naturalness by continuously generating visible behavior during non-speaking intervals. In the idle state, the agent does not collapse into a frozen portrait; it maintains identity, gaze, posture, breathing, and subtle facial motion over the streaming history. In the listening state, the model can produce responsive non-verbal feedback such as gaze shifts, nods, micro-expressions, and posture changes that are temporally coupled with the user's speech and visual cues. Since speech and video latents are predicted from the same causal context before decoding, lip motion, facial dynamics, and prosody are synchronized natively rather than repaired by post-hoc alignment. These properties make the generated agent feel closer to a real interlocutor: it is visually present while waiting, attentive while the user speaks, and coherent when transitioning between listening, thinking, and speaking.

\textbf{Interruption and proactive speaking.}
The full-duplex behavior of \model is learned from interleaved interaction data instead of being implemented only as hand-crafted turn-taking rules. During training, user inputs and agent outputs from text, audio, and video are placed on the same causal timeline, so the model observes when humans continue, pause, overlap, interrupt, yield, or resume. At inference time, the model keeps consuming user audio-video observations even while generating its own response, allowing it to stop, shorten, or redirect its speech when the user naturally interrupts. The same unified context also enables proactive speaking: when salient visual events, objects, expressions, or user actions appear in the input stream, the model can initiate a relevant comment or question based on what it sees, rather than waiting for an explicit spoken request. This turns interaction from a passive question-answer loop into a more human-like continuous exchange.

\FloatBarrier

\section{Related Works}

\textbf{Full-duplex spoken dialogue.}
Recent spoken dialogue models move beyond turn-based speech pipelines by modeling listening and speaking on a shared streaming timeline~\citep{chen2025fullduplexsurvey}. Moshi models user and assistant speech as parallel streams and removes explicit turn segmentation~\citep{defossez2024moshi}; OmniFlatten adapts a GPT backbone into an end-to-end full-duplex speech-text dialogue model~\citep{zhang2024omniflatten}; SALM-Duplex directly models continuous user speech inputs and codec-based assistant outputs~\citep{hu2025salmduplex}; and DuplexSLA further adds a synchronized action stream for planning and tool use during speech~\citep{zhang2026duplexsla}. Commercial systems such as Seeduplex have also reported native end-to-end full-duplex speech interaction with always-on listening and interference suppression~\citep{seeduplex2026}. These works establish that real-time dialogue should not be formulated as alternating ASR--LLM--TTS turns. However, they are primarily speech or speech-text systems: they do not generate a visual agent, do not consume streaming video observations, and therefore cannot learn visual listening behavior, facial/body feedback, or audio-visual response timing as part of the same model.

\textbf{Interactive digital humans and audio-visual avatars.}
Audio-driven avatar generation has progressed from portrait and talking-head synthesis toward real-time full-body and long-duration character animation~\citep{xu2024vasa,lin2025omnihuman1,ding2025kling-avatar}. Streaming methods such as TalkingMachines, StreamAvatar, LiveTalk, Hallo-Live, and OmniForcing improve latency and temporal consistency for audio-visual generation~\citep{talkingmachines2025,streamavatar2025,livetalk2025,hallolive2026,omniforcing2026}. Recent interactive digital-human systems further connect multimodal understanding or dialogue modules with real-time visual generation, including MIDAS, MAViD, M.I.O, U-Mind, LPM, and X-Streamer~\citep{chen2025midas,mavid2025,mio2025,deng2026umind,lpm2026,xstreamer2025}. These systems are important steps toward visual conversational agents, but most of them still assemble separate modules: a speech or language model decides what to say, while an audio-driven or instruction-driven visual generator renders the character. As a result, duplex behavior is often handled by system-level control, VAD, turn-taking logic, or external audio models, rather than learned end to end together with visual perception and visual expression.

\textbf{End-to-end audio-visual interaction.}
A smaller set of works has explored more unified audio-visual agents. Body of Her is a notable preliminary study of an end-to-end humanoid agent that integrates audio and visual inputs and models speech, full-body behavior, idling, response, and manipulation in real time~\citep{ao2024bodyofher}. FlowAct-R1 and related humanoid video models also point toward action-level interactive generation~\citep{flowact2026}, while causal video and world-modeling methods provide tools for streaming rollouts and long-horizon consistency~\citep{chen2024diffusionforcing,huang2025selfforcing,liu2025rollingforcing,worldplay2025,matrixgame2026}. Wan-Streamer follows this end-to-end direction but targets a stricter setting: language, audio, and video appear on both the input and output sides, are modeled by a single Transformer, and are served with fully causal encoders, decoders, and full-history streaming inference. This allows perception, reasoning, speaking, visible listening, interruption handling, and synchronized audio-visual generation to be learned as one native full-duplex process rather than connected as a cascade.

\section{Conclusion}
We presented \model, a native-streaming, end-to-end foundation model for real-time full-duplex text, audio, and video interaction. Unlike cascaded systems that alternate among perception, language modeling, speech synthesis, and visual generation modules, \model represents user inputs and agent outputs across all modalities as one causal stream processed by a single Transformer. With fully causal audio and video VAEs, causal encoders and decoders, and a block-causal Transformer, the model can perceive current observations, generate synchronized audio-visual responses, emit each streaming unit, and commit the generated latents back into history with minimal delay. Together with the thinker-performer serving design, \model reaches sub-second interactive latency while preserving full-history context. The current v0.1 results are validated at a preliminary 192p output resolution; this serves as a proof of concept for the end-to-end streaming design, and scaling to higher resolutions is straightforward and left to future work. These results suggest that real-time multimodal agents should be designed from the ground up as native full-duplex systems, where listening, seeing, speaking, and visible response are learned jointly rather than assembled as post-hoc modules.

\clearpage
\newpage
\bibliographystyle{style/plainnat}
\bibliography{paper}

\clearpage
\newpage

\appendix
\section*{Appendix}

\section{Contributions and Acknowledgements}
\label{app:contributions-and-acknowledgements}

\subsection{Core Contributors}
Lianghua Huang, Zhi-Fan Wu, Wei Wang, Yupeng Shi, Mengyang Feng, Junjie He, Chen-Wei Xie, Yu Liu, and Jingren Zhou.

\subsection{Contributors}
Contributors are listed alphabetically by first name: Ang Wang, Bang Zhang, Baole Ai, Chen Liang, Cheng Yu, Chongyang Zhong, Jinwei Qi, Kai Zhu, Pandeng Li, Peng Zhang, Wenyuan Zhang, Xinhua Cheng, Yitong Huang, Yun Zheng, Yuzheng Wang, and Zoubin Bi.

\end{document}